\documentclass[10pt,twocolumn,letterpaper]{article}

\usepackage{wacv}
\usepackage{times}
\usepackage{epsfig}
\usepackage{graphicx}
\usepackage{amsmath}
\usepackage{amssymb}

\usepackage{multirow}

\usepackage[pagebackref=true,breaklinks=true,letterpaper=true,colorlinks,bookmarks=false]{hyperref}

\wacvfinalcopy 

\def\wacvPaperID{400} 

\ifwacvfinal\fi
\begin{document}

\title{Learning On-Road Visual Control for Self-Driving Vehicles with Auxiliary Tasks}


\author{
Yilun Chen\textsuperscript{1}, Praveen Palanisamy\textsuperscript{2}, Priyantha Mudalige\textsuperscript{2},  Katharina Muelling\textsuperscript{1}, John M. Dolan\textsuperscript{1}\\
\textsuperscript{1}The Robotics Institute, Carnegie Mellon University\\
\textsuperscript{2}General Motors Global R\&D \\
\tt\small \{yilunc1, jdolan\}@andrew.cmu.edu, kmuelling@nrec.ri.cmu.edu, \\
\tt\small \{praveen.palanisamy, priyantha.mudalige\}@gm.com
}

\maketitle
\ifwacvfinal\pagestyle{empty}\fi

\begin{abstract}
A safe and robust on-road navigation system is a crucial component of achieving fully automated vehicles. NVIDIA recently proposed an End-to-End algorithm that can directly learn steering commands from raw pixels of a front camera by using one convolutional neural network. In this paper, we leverage auxiliary information aside from raw images and design a novel network structure, called Auxiliary Task Network (ATN), to help boost the driving performance while maintaining the advantage of minimal training data and an End-to-End training method. In this network, we introduce human prior knowledge into vehicle navigation by transferring features from image recognition tasks. Image semantic segmentation is applied as an auxiliary task for navigation. We consider temporal information by introducing an LSTM module and optical flow to the network. Finally, we combine vehicle kinematics with a sensor fusion step. We discuss the benefits of our method over state-of-the-art visual navigation methods both in the \textit{Udacity} simulation environment and on the real-world \textit{Comma.ai} dataset. 
\end{abstract}

\section{Introduction}
Perception and control have long been two related key challenges researched separately in the autonomous driving industry. Recent advances in deep learning have introduced End-to-End learning as a new method for learning driving policies for self-driving cars. Unlike traditional approaches \cite{wei2013towards} that divide the system into two separate perception and control parts which contain tasks like lane detection, path planning and control logic, End-to-End approaches often directly learn the mapping from raw pixels to vehicle actuation. Recent demonstrations have shown some successful examples of training systems End-to-End to perform simple tasks like lane-keeping \cite{chen2017end} or obstacle avoidance \cite{muller2006off}.

\begin{figure}[t]
 \centering
 \includegraphics[width=\linewidth]{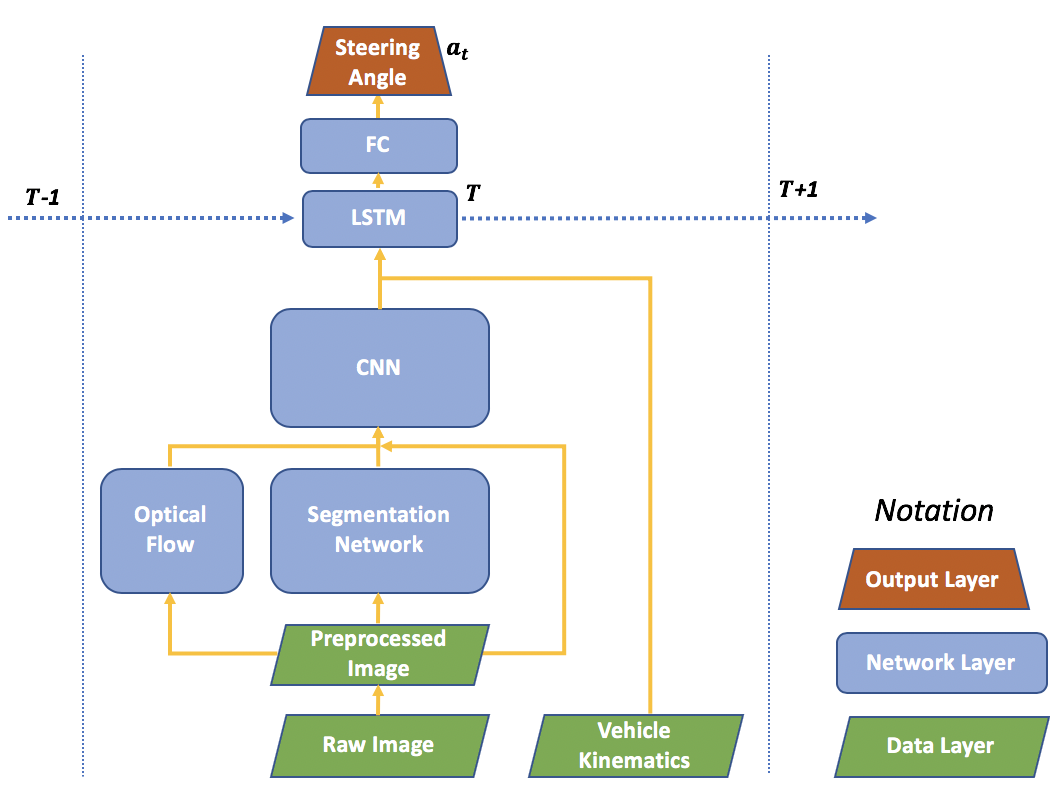}
 \caption{The proposed End-to-End learning architecture: Auxiliary Task Network (ATN). The ATN network considers the auxiliary task of Image Segmentation, Transfer Learning and Optical Flow, and auxiliary feature input from LSTM and Vehicle Kinematics.}
 \label{fig-overview}
\end{figure}

The End-to-End methods have the advantage of direct optimization without manually defined rules, which results in better performance and less human engineering effort. However, current End-to-End models use a single deep convolutional neural network to map from perception to control \cite{nvidiacar}. This straightforward representation simplifies the modeling but has a number of drawbacks. First, it loses a lot of detailed dynamic information for the specific driving task. Speed, map information and vehicle mechanics are all important for the driving task. Second, it lacks human prior knowledge of the network. People obey a lot of basic assumptions like traffic rules when driving, whereas the network is expected to learn from scratch. Third, it only considers current perception information for decision making. In contrast, people memorize history information and process it along with current perception information to drive.

To address these problems, we present an improved learning approach with auxiliary tasks. We enrich the algorithm with dynamic information by performing an extra image segmentation task and introducing vehicle kinematics. Human domain knowledge is introduced by adopting transfer learning from the large-scale image recognition task. The temporal information is further included by using an Long Short Term Memory (LSTM) network \cite{hochreiter1997long} and optical flow. The new approach is built on top of a convolutional neural network, with separate components to perform auxiliary tasks. Our proposed network structure is shown in Fig. \ref{fig-overview}. 

Our system differs from the original network structure in four respects: (1) the system takes advantage of additional information by first obtaining a segmentation map instead of directly using the implicit raw image; (2) our learning system transfers knowledge from other tasks and hence speeds up training; (3) we consider the temporal information by adding a recurrent module into the network and further introducing optical flow as an auxiliary task; (4) we use vehicle kinematics information in addition to image input. The overall system uses auxiliary tasks to guide more efficient training but still maintains the benefit of being trained End-to-End. The resulting architecture can lead to faster convergence and higher accuracy in performance. 

\section{Related Work}

Deep neural networks have been proven to be very successful in many fields. Recently a lot of work focuses on applying deep networks to learn driving policies from human demonstrations. One of the earliest attempts originates from ALVINN \cite{pomerleau1989alvinn}, which used a neural network to directly map front-view camera images to steering angle. NVIDIA \cite{nvidiacar} recently extended this approach with deep neural networks to demonstrate lane following in more road scenarios. Other concurrent examples of learning End-to-End control of self-driving vehicles include \cite{chen2017end, rausch2017learning}. These works emphasize an End-to-End learning procedure by using a single deep network to learn driving policy and not requiring further architecture design or human prior knowledge. However, this has resulted in problems of data inefficiency and bottlenecks in learning more complex driving behavior.

One way to overcome these problems is to train on a larger dataset. Xu et al. \cite{xu2017end} scaled this effort to a larger crowd-sourced dataset and proposed the FCN-LSTM architecture to derive a generic driving model. Another way is to set intermediate goals for the self-driving problem. Chen et al. \cite{chen2015deepdriving} and Al-Qizwini et al. \cite{al2017deep} map images to a number of key perception indicators, which they call affordance. The affordance is later associated with actions by hand-designed rules. Some other approaches include applying an attention model in learning to drive \cite{chen2017brain, kim2017interpretable} and using hierarchical structures to learn meta-driving policies \cite{liaw2017composing}.

In this paper, we decide to improve learning driving policies by adding auxiliary tasks. The idea of using auxiliary tasks to aid the nominal task is not unprecedented. Max et al. \cite{jaderberg2016reinforcement} trained a reinforcement learning algorithm with unsupervised auxiliary tasks. By forecasting pixel changes and predicting rewards, the reinforcement learning algorithm converges faster and to a higher reward. The concept of transfer learning \cite{pan2010survey} can also be regarded as training the nominal task with auxiliary tasks. The benefit of training with auxiliary tasks can be faster convergence time and better performance.

In determining the auxiliary tasks we need, we should consider both efficiency and importance. Efficiency means we can easily integrate the auxiliary task into the network without breaking the End-to-End training. Importance means the task must address one of the drawbacks in the existing system. Luckily, there are numerous tasks related to autonomous driving, but they are seldom combined together. Image segmentation \cite{badrinarayanan2017segnet} helps derive better cognition of the environment. Optical Flow \cite{fortun2015optical} can recognize the movement of objects. Transfer learning \cite{pan2010survey} uses the power of sharing common features across tasks. Finally, LSTM \cite{hochreiter1997long} can be used to extract temporal information. We propose to unify all these efforts and combine them into one network and workflow to learn the driving policy.

\section{Method}

The Auxiliary Task Network, as shown in Fig.\ref{fig-overview}, is built upon the single convolutional network architecture \cite{nvidiacar} for self-driving vehicles. We introduce several independent auxiliary tasks into the original network to boost the performance. The auxiliary tasks we introduce here include incorporating image segmentation, transferring from existing models, using a recurrent module, adopting optical flow and adding vehicle information. The final model is a combination of different component networks.

The high-level data flow through the network can be described as follows. The input data to the network are raw RGB images with a front view of the autonomous vehicle. First, we do image preprocessing to calibrate images and increase the diversity of data. The preprocessed data are then used by two modules in parallel. The first module feeds preprocessed data into a pretrained segmentation network to obtain the corresponding segmentation map. The second module constructs a deep neural network to estimate optical flow. Next, a traditional convolutional network is used to extract spatial features from an augmentation of the segmentation map, optical flow map, and preprocessed image. We further concatenate additional vehicle kinematics with the context features from the convolutional network to obtain the context features space. We then adopt an LSTM network to process the features and incorporate the temporal dynamics of the system. This enables the system to memorize the past observations. Finally, the LSTM output is fed into a fully connected layer to learn the continuous steering angle output.

\subsection{Auxiliary Image Segmentation}
Image segmentation has been widely researched for decades. In autonomous driving, image segmentation is often performed to classify and understand the surrounding environment of the ego vehicle, for example, to categorize surrounding vehicles, pedestrians, road boundaries, buildings, etc. 

The image segmentation result contains a large amount of information since it performs classification with pixel-level precision. However, current image segmentation is hard to directly apply in the autonomy stack of a self-driving vehicle. Industry prefers to use object detection for classification purposes, as it gives the position in a more straightforward way. 

In this paper, we propose to use image segmentation in a novel way. Instead of treating image segmentation as a separate perception task, we train and perform image segmentation as an auxiliary task of learning for control. The learned segmentation map is augmented with the original image to derive better features for the control policy. We believe the learned segmentation map contains much richer information for controlling the car's behavior. For example, the segmentation map can explicitly identify where the road boundary is and where the surrounding vehicles on the road are. This can make the vehicle learn its stop/go behavior or turning behavior much more easily. It significantly decreases the difficulty of learning everything implicitly from the original preprocessed image.

\begin{figure}[t]
 \centering
 \includegraphics[width=\linewidth]{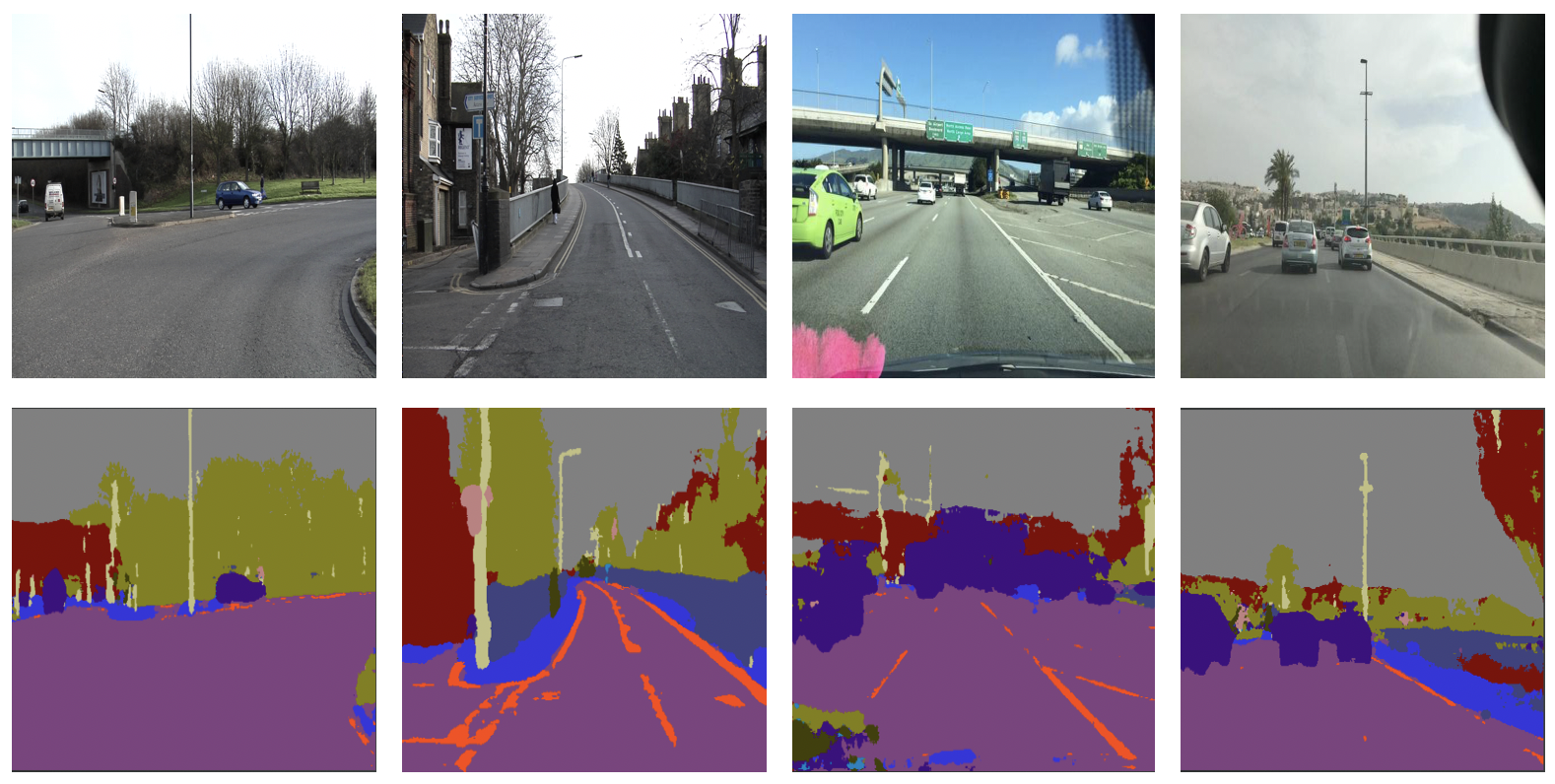}
 \caption{Examples of segmentation performed on the preprocessed images in \textit{Comma.ai} dataset. }
 \label{fig:segmentation}
\end{figure}

Another benefit of using image segmentation as an auxiliary task is that we can obtain this capability offline without training extra data. We can apply public image segmentation models trained on a much larger dataset and only fine-tune the model with minimal data. In the implementation, we have segmentation categories of the sky, road, lane markings, building, traffic lights, pedestrian, tree, pavement, and vehicle. The preprocessed images are passed into a segmentation network and the output segmentation map will be stacked into the depth channel with the original preprocessed images and optical flow map before being fed into the Convectional Neural Network (CNN) module. Examples of segmentation masks used are shown in Fig.\ref{fig:segmentation}.

\subsection{Transferring from Existing Tasks}
\begin{figure}[t]
 \centering
 \includegraphics[width=\linewidth]{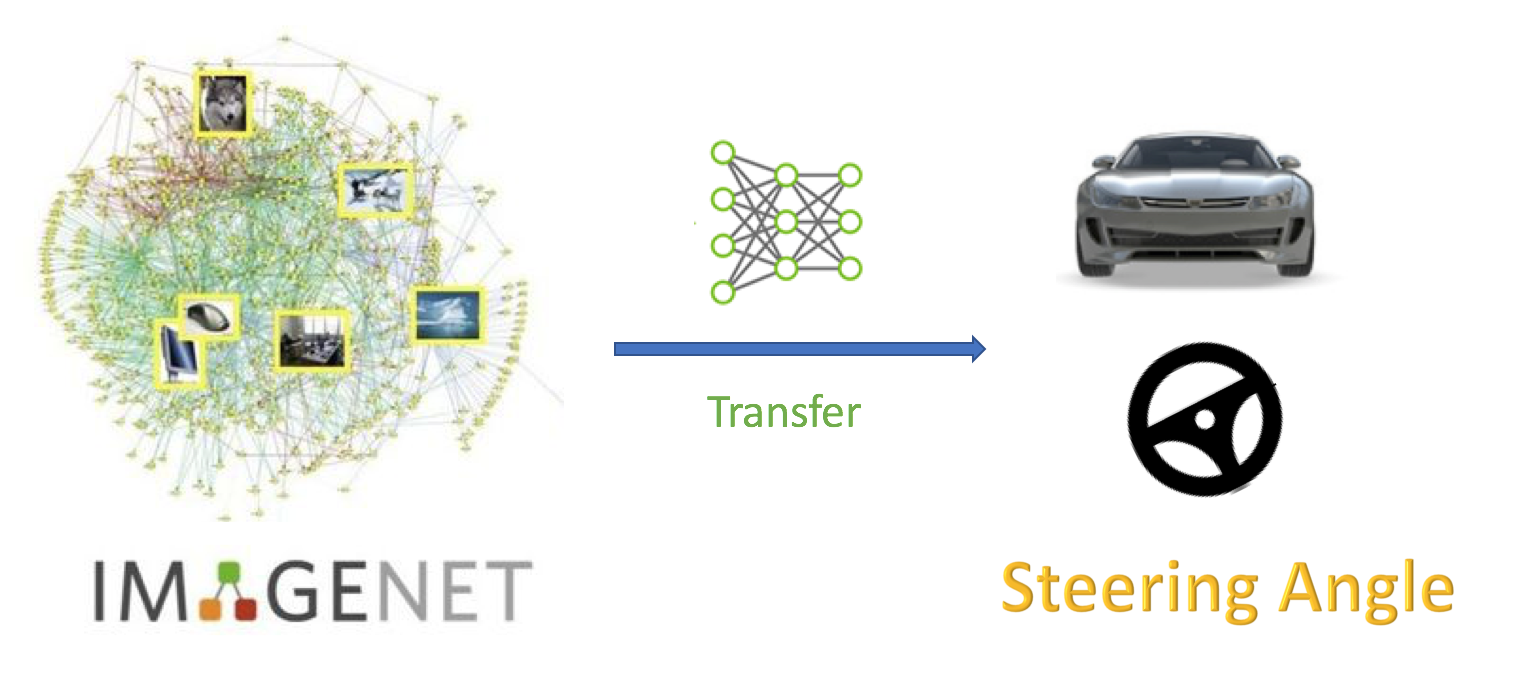}
 \caption{Illustration of transfer learning from object recognition in Imagenet to learning steering angle for the self-driving car.}
 \label{fig-transfer}
\end{figure}

CNN has recently been applied to a number of common and practical tasks and proven to be very successful. The most popular task is to learn object recognition on the Imagenet dataset that contains 1.2 million images of approximately 1000 different classes \cite{krizhevsky2012imagenet}. Currently, numerous famous network structures in the literature have been proven to be powerful. The best-performance trained model can learn a generic set of features, and recognize a large variety of objects with high accuracy. The intermediate features learned are found to have universal expressiveness across multiple domains. This motivates us to leverage the power of pre-trained networks and apply transfer learning \cite{pan2010survey} from object classification to learning driving policy, as shown in Fig.\ref{fig-transfer}.

In this paper, we compare three models: the ResNet \cite{he2016deep}, the VGG network \cite{simonyan2014very} and baseline CNN for the CNN module in the network. The ResNet and VGG networks are the champions of Imagenet Challenge in 2016 and 2014, perfectly trained and optimized for Imagenet. The baseline CNN is the same from NVIDIA self-driving vehicle \cite{nvidiacar}. For feature extraction purposes, we only choose the convolutional layers of the whole network. The ResNet and VGG net will be used as pre-trained models and fine-tuned during training, while the baseline CNN will be trained totally from scratch. We prune the network to make sure each network has approximately the same number of parameters.

\subsection{Temporal Information}
Decision making for an autonomous vehicle is not an independent choice at every time step. A human would consider past environment information and previous actions taken and then make a consistent decision of driving behavior. This requires the system to not only rely on the current state but also incorporate past states. To deal with temporal dependencies, we propose two solutions: adding a recurrence module and using optical flow. 

\subsubsection{Recurrence Module}
After concatenating spatial features from the CNN layer with vehicle kinematics, an LSTM layer is added to process previous information of the environment. The LSTM processes the context feature vector $v$ in a sliding window of size $w$. This implies the steering action prediction result is dependent on $w$ past input observations $X_{t-w+1}-X_t$. By changing the parameter of $w$, we can alter how long the system takes to make the decision. Small $w$ leads to shorter-term memory, so it has faster reaction time but is prone to sudden sensor failure. Larger $w$, on the other hand, leads to a much smoother and more stable behavior. The downside of larger $w$ is that it requires longer training and test time for choosing actions. The LSTM fuses $w$ past states and current state and then learns a temporal-aware model. This representation can alleviate the problem brought by the Partially Observable Markov Decision Process (POMDP) environment \cite{spaan2012partially}.

\subsubsection{Optical Flow}
\begin{figure}[t]
 \centering
 \includegraphics[width=\linewidth]{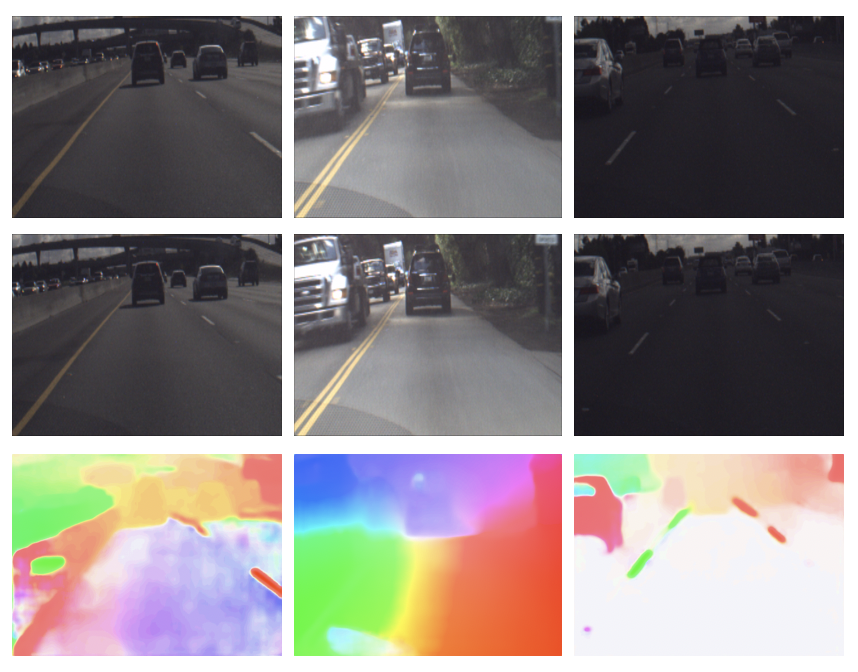}
 \caption{Examples of Optical flow performed on the preprocessed images in \textit{Comma.ai} dataset. }
 \label{fig-opticalflow}
\end{figure}
Optical flow is a traditional vision task to estimate the relative movements of objects when observation changes. When applied between two successive images through time, optical flow can be used to evaluate the change of environment so as to infer our own movement.
The task of learning steering angles can use the help of optical flow to construct connections to the changing environment. As the image in a certain frame only contains static environment information, an optical flow map can serve as a good supplement of dynamic environment information.

Recent advances have applied CNN to estimate optical flow as a supervised learning task \cite{dosovitskiy2015flownet}. This provides the chance to combine optical flow as part of our network serving as an auxiliary task. The optical flow map is generated against two successive preprocessed images and then stacked into the depth channel with the original preprocessed images and segmentation map before feeding into the CNN module. Examples of optical flow output are shown in Fig.\ref{fig-opticalflow}.

\subsection{Additional Vehicle Information}
We further hypothesize that visual input alone is not enough to make a good steering angle decision. The vehicle's behavior can be better estimated by adding the vehicle's kinematic information. The kinematic information ensures that the car does not execute a driving behavior that is against some physical rules.

It can be speculated that making a U-turn at 10 mph and 30 mph is different regarding the turning angle and the strategy used. However, the visual observations given are almost the same: although we can infer the speed of the vehicle by the scene change speed, it remains ambiguous and not easy to learn from images. The vehicle kinematics can provide information such as current vehicle speed.

Limited to the simulation environment and real-world dataset, we select the kinematic parameters as an extra input to the LSTM layer, shown in Table.\ref{tab:kinematics}.

\begin{table}[t]
    \centering
    \bgroup
    \def\arraystretch{1.2}
    \begin{tabular}{r|l}
        Vehicle Kinematics & Value Range\\
        \hline
        acceleration rate & $[0,5] m/s$ \\
        speed & $[0,60] mph$\\
        heading & $[-45, 45]^{\circ}$ \\
        distance to road curb & $[-2,2] m$ \\
        previous steering angle & $[-45, 45]^{\circ}$
    \end{tabular}
    \caption{Vehicle Kinematics used as additional vehicle information to the network.}
    \label{tab:kinematics}
    \egroup
\end{table}

\section{Experiment Setup}
Unlike traditional vision tasks that have explicit ground truth to refer to, the target we are learning is a control policy. Although a supervised learning approach is used to tackle this problem, our experiments and evaluation should not be limited to how precisely the algorithm can recover the same human reaction on a given dataset. It should be taken into account how the learned model can actually react in the real world scenarios. To this end, we evaluate our algorithm in two different ways: on a real-world driving dataset and a simulation platform.

\subsection{Experiment environment}
There are currently numerous public datasets that provide front-view images of on-road driving, but few of them accompany images with the corresponding control commands of the human driver. Among the available datasets, the \textit{Comma.ai} dataset provides a set of high-quality consistent driving data on the highway scenario. The \textit{Comma.ai} dataset contains 7.5 hours of highway driving. The sensor input is recorded at 20 Hz with a camera mounted on the windshield of an Acura ILX 2016. Together with the images, the dataset contains information such as the car's speed, acceleration, steering angle, GPS coordinates, and gyroscope angles.

The simulation environment we choose is the public open-source \textit{Udacity} simulator. The \textit{Udacity} simulator is an open-source simulator developed based on \textit{Unity}. The simulation gives a realistic 3D visualization of the vehicle driving on three given tracks in the desert, mountain, and forest. It supports easy recording of a human driver's actions and we can have access to all the driver's control actions together with sensory perception information.
 
\subsection{Data Preprocessing}
\begin{figure*} [!htb]
\centering
\includegraphics[width=0.9\linewidth]{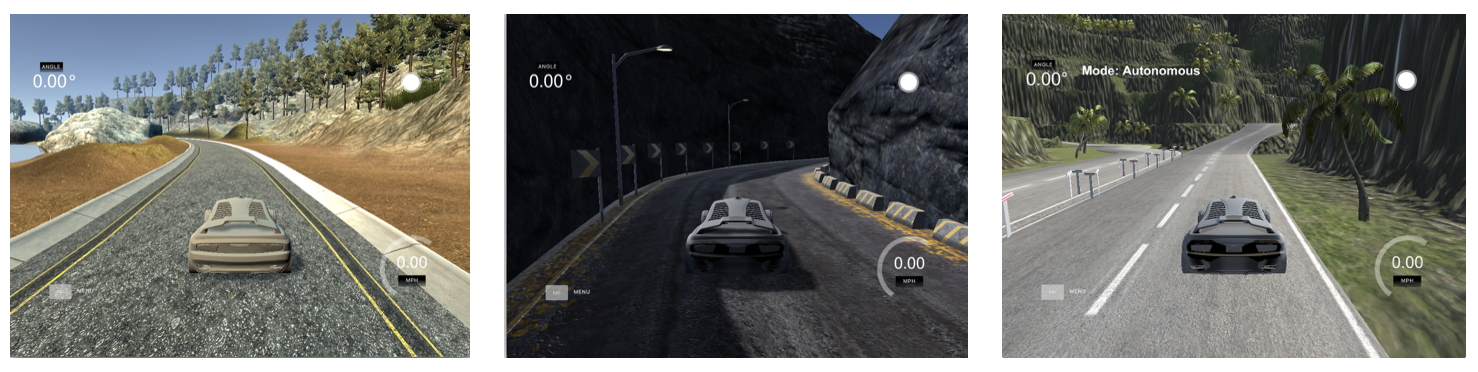}
\caption{Sample screenshots of the environment in the \textit{Udacity} autonomous driving simulator. The left one shows the training track in the desert, while the two on the right show the test track in mountain and suburb. The test sets are different from the training set regarding lighting conditions, curvature, inclination, road geometry, and off-road scenery and thus are considered much more difficult.}
\label{fig:simulation}
\end{figure*}

In the \textit{Udacity} simulation environment, we use three tracks. The three tracks severally depict a highway running through a desert, suburb, and mountain. Example screenshots of the different trials are shown in Fig. \ref{fig:simulation}. The desert track is used for training purposes, and the suburb and mountain tracks are used for testing. 

We collected image data for training by driving the car in the simulation environment. To introduce various driving styles from multiple people, we collected data from six people, each driving the desert track twice. We recorded the steering angle, speed, acceleration rate and braking rate paired with the corresponding images while driving with keyboard input. The system operates at a 10-hertz frequency. Altogether we collected 6245 images, totaling about 1 hour of driving data. We sampled images at 2 Hz to prevent redundant pictures. The images captured simulate the front view from the vehicle via a camera mounted on top of the car.

The images obtained are not directly used for training purposes. Before training, we preprocess and augment the data using techniques similar to those described in \cite{nvidiacar}. Data augmentation is used here to increase the size of the training set and also the diversity of training samples. The following operations were performed.

\begin{itemize}
\item \textbf{Cropping}: The images are cropped to remove extraneous elements. We removed the top of an image which includes a large area of sky or mountain tops and the bottom of the image which contains the car hood.
\item \textbf{Upsampling}: In the original training set, most scenarios are going straight along the road. Images with a steering angle larger than 15 degrees are scarce compared to a significant number of training samples with a steering angle less than 5 degrees, which means the steering angle distribution is strongly biased towards zero. To overcome the problem of imbalanced data, we manually upsample images with steering angle greater than 10 degrees by ten times and steering angle greater than 5 degrees by five times. The data are then randomly shuffled before training.
\item \textbf{Brightness}: Each frame is first converted to HSV space and the value channel is then multiplied by a random value from 0 to 10. This changes the lighting situation and makes the model adaptive to all kinds of weather.
\item \textbf{Flipping}: We flip all frames to obtain their mirror in the horizontal direction. This helps to make sure we have exactly the same amount of left and right turning samples. This is to prevent the algorithm from suffering from any bias in the left or right direction.
\end{itemize}

For the real data from the \textit{Comma.ai} driving dataset, there is no need for cropping and brightness preprocessing. We use the same upsampling and flipping techniques to deal with the data balance problem. The dataset includes 11 video clips of 7.5 hours of highway driving at 20 Hz. Here we only want to consider stable highway driving at normal speed in daylight. We further exclude the driving videos at night or in traffic jams with speed under 10mph. The finally selected footage has a length of about 2 hours of driving. We split it by using 130K frames for training and 20K frames for testing.

\subsection{Implementation Detail}

As a baseline to compare our algorithms, we implement the CNN network structure proposed in \cite{nvidiacar}. The difference in our implementation is that we add batch normalization \cite{ioffe2015batch} and dropout \cite{tinto1975dropout} layers after each convolutional layer for better convergence and performance. The same is done on the VGG net and ResNet structure. The baseline CNN network consists of 5 convolutional layers. The first three layers have a kernel size of $5 \times 5$, and the last two layers have a kernel size of $3 \times 3$. The depth of each feature map is 24, 36, 48, 64, 64. The activation function we use here is ReLu \cite{nair2010rectified}.

For the optical flow task, we use the implementation of FlowNetSimple \cite{dosovitskiy2015flownet}. For the image segmentation task, we use the implementation of SegNet from \cite{badrinarayanan2017segnet}. For both of these auxiliary tasks, we train the network using the public benchmark dataset and then fine-tune the network with our domain dataset.

The LSTM layer we use follows the implementation in \cite{zaremba2014recurrent}. The LSTM has 128 hidden units. We conducted grid-search for the optimal window size $w$ for the LSTM and found that $w=3$ generates the best result, which means we look back for 1.5 seconds. The FCN module consists of 3 fully connected layers with 256, 128 and 64 units followed by a $Tanh$ activation function. The steering angle output is a continuous value with a range of $[-90,90]$ degree.

We use Adam optimization \cite{kingma2014adam} to train all networks. The learning rate starts off at 0.001 with a momentum decay of 0.9. Every time the training curve plateaus, we decrease the learning rate by half.

\section{Evaluation and Results}
We evaluate the effectiveness of our proposed methods using auxiliary tasks from two perspectives: how close is our policy actions from human behavior and how well can our agent perform in real scenarios. To this end, we define three metrics: Root Mean Square Error, Mean Continuity Error and failure per 10 km. The evaluations are performed both on a simulated environment \textit{Udacity} and a real-world dataset \textit{Comma.ai}.
\subsection{Evaluation Metrics}
We report our experimental results on the \textit{Udacity} simulation and the \textit{Comma.ai} dataset. For the steering angle prediction task, Root Mean Square Error (RMSE) is used as the evaluation metric. RMSE expresses the average system prediction error on the dataset. The RMSE metric is defined as
\begin{equation*}
RMSE = \sqrt{\frac{1}{|D|}\sum_{i=1}^{|D|}(\hat{a}_i - a_i)^2},
\end{equation*}
where $\hat{a}_i$ and $a_i$ are the ground-truth and predicted steering angle in degress for frame $i$. $|D|$ is the total number of frames in the test set.

The RMSE can estimate the precision of the system but can neglect the stability of the prediction. To address stability, a stability metric based on the fluctuation of our prediction is proposed. The intuition is to encourage the predictions to change smoothly without any sudden jump in two successive steering angle predictions. We call this metric Mean Continuity Error (MCE), which is defined as

\begin{equation*}
MCE = \sqrt{\frac{1}{|D|-1} \sum_{i=1}^{|D|-1} (a_{i+1} - a_i)^2}
\end{equation*}
where $a_i$ and $a_{i+1}$ are the predicted steering angle in degrees for frame $i$ and $i+1$ respectively. $|D|$ is the total number of frames in the test set.
The MCE metric reflects the consistency of the algorithm's prediction through time, which is valuable in terms of evaluation of the consistency of the learned driving policy.

The RMSE and MCE metrics mainly evaluate how well we can mimic human driving behavior. In addition, we prefer to directly evaluate how the learned agent can react in a real driving environment. So we count the number of times the vehicle fails when applying the learned driving policy. In the simulation environment, a failure is triggered when we have a lane boundary violation. In the real world dataset, the situation is more complex, as the environment cannot change with respect to our choice of actions and we could never generate a real failure. In this case, we treat 10 continuous predictions (1 second's worth) with a deviation greater than 5 degrees from human driving data as a failure.

\subsection{Comparison of Auxiliary Tasks}
\begin{table*}[t]
\centering
\begin{tabular}{ |c|c c c|c c c| } 
\hline
\multirow{2}{10em}{Networks with Auxiliary Tasks} & \multicolumn{3}{|c|}{\textit{Udacity} Simulation} & \multicolumn{3}{|c|}{\textit{Comma.ai} Dataset} \\
\cline{2-7}
& RMSE & MCE & Fail/10km & RMSE & MCE & Fail/10km \\
\hline
\hline
\cite{nvidiacar} NVIDIA. & 1.78 / 2.67 & 0.53 / 0.68 & 1.2 / 5.0 / 12.3 & 4.67 / 7.34 & 2.01 / 2.33 & 18.3 \\
\hline
ATN w/o Image Segmentation & 1.49 / 2.26 & 0.44 / 0.50 & 0.4 / 3.2 / 7.6 & 3.95 / 6.01 & 1.66 / 1.92 & 10.9 \\
ATN w/o LSTM & 1.09 / 1.44 & 0.37 / 0.45 & 0 / 2.1 / 3.8 & 2.89 / 3.95 & 1.01 / 1.20 & 5.6  \\
ATN w/o Optical Flow & 1.03 / 1.30 & 0.26 / 0.37 & 0 / 1.6 / 3.3 & 2.75 / 3.72 & 0.92 / 1.08 & 4.9 \\
ATN w/o Vehicle Kinematic & 1.13 / 1.32 & 0.33 / 0.45 & 0 / 1.9 / 4.1 & 3.04 / 4.45 & 1.05 / 1.31 & 5.2 \\
ATN w/ base & 0.98 / 1.23 & 0.23 / 0.32 & 0 / 1.1 / 2.6 & 2.56 / 3.43 & 0.82 / 0.97 & 4.3 \\
\hline
ATN w/ VGG & 0.92 / 1.13 & 0.21 / 0.29 & 0 / 1.0 / 2.4 & 2.37 / 3.24 & 0.71 / 0.82 & 3.7  \\
ATN w/ Res & \textbf{0.88} / \textbf{1.09} & \textbf{0.20} / \textbf{0.27} & \textbf{0} / \textbf{0.9} / \textbf{2.3} &  \textbf{2.23} / \textbf{3.12} & \textbf{0.65} / \textbf{0.77} & \textbf{3.4} \\
\hline
\end{tabular}
\caption{Comparison between different network structures for predicting steering angles in the \textit{Udacity} Simulation and \textit{Comma.ai} Dataset. Auxiliary Task Network (ATN) refers to the final network structure with all auxiliary tasks, as shown in Fig.\ref{fig-overview}. We evaluate all network performance against 3 metrics: RMSE, MCE and Fails per 10km.}
\label{tab-metrics}
\end{table*}
We evaluate the influence of each auxiliary task and auxiliary feature input with a comparison to the baseline method (see Table.\ref{tab-metrics}). Row 2-6 in Table.\ref{tab-metrics} shows the comparison with and without 4 auxiliary tasks or auxiliary feature input in the ATN network (Fig.\ref{fig-overview}): Image Segmentation, Recurrence (LSTM), Optical Flow and Vehicle Kinematics. We can see each auxiliary task or feature input has contributed to the ATN network. Without any of the 4 auxiliary tasks, the performance of ATN will drop. The two columns in RMSE and MCE show the training and testing score of each metric. We can observe that the Image Segmentation task contributes most to the ATN network, as the RMSE increases from 1.23 degrees to 2.26 degrees and MCE increases from 0.32 degrees to 0.50 degrees in the simulation. The Optical Flow task has the least influence on steering angle prediction, since the loss in RMSE and MCE only increase from 1.23 to 1.30 and 0.23 to 0.26 in the simulation without Optical Flow. 

In addition to the quantitative results, we also investigate specific examples that make the performance better. With image segmentation, there are much fewer cases where the vehicle runs out of road boundary. Many of the previous cases where the vehicle gets confused because of the ambiguity of road and sidewalk are now solved with the direct guidance from image segmentation result. With LSTM, we observe that absurd outlier predictions with deviation more than 10 degrees completely disappear. The cases where two sequential frames make dramatically different steering angle predictions also drop significantly, which adds great stability to the algorithm. With Vehicle Kinematics, the overall driving behavior becomes more consistent with less obvious oscillation on the road.

The Failure cases test also supports our findings. The three columns in Fail in simulation correspond to the three environments in the \textit{Udacity} simulation: desert, suburb and mountain. We use the desert environment as the training set and use the suburb and mountain environments as the test set. As is shown, driving in the desert environment can be fully solved as we get almost zero failures over the 10km distance. The failures happen more often in the suburb and mountain environments, with a final failure frequency of 1.1 and 2.6 times per 10km due to the fact that the scenery is unseen and has greater difficulty. The comparison between with and without auxiliary tasks aligns with our findings in RSME and MCE.

From row 6-9, we can compare the performance changes with and without transferring features from the Imagenet Recognition task. The ``base" CNN are totally trained from scratch while the ``VGG" and``Res" CNN are networks with parameters pretrained for the Imagenet task. By using the transferred features from the VGG network and ResNet, the RMSE test score in simulation drops down from 1.23 to 1.17 and 1.09 respectively. This implies the features from the image recognition task are universal and transferable to the task of learning the steering angle.

Finally, we compare the best-performance ATN using ResNet with the baseline NVIDIA \cite{nvidiacar} method (row 1 and 8 in Tab. \ref{tab-metrics}). We can see the test RMSE drops from 2.67 to 1.09 with 59.5 percent in the \textit{Udacity} simulation and from 7.34 to 3.12 with 57.5 percent in the \textit{Comma.ai} dataset. The failures frequency per 10km drops from 5.0 and 12.3 to 0.9 and 2.3 in the suburb and mountain environments in the \textit{Udacity} simulation, and from 18.3 to 3.4 in the \textit{Comma.ai} dataset. As can be observed, the performance improvement is huge with respect to both similarity to human behavior and driving capability.

\subsection{Discussion on Environments}
In the experiment, we test both in a simulation environment and on a real-world driving dataset. From the metrics Table \ref{tab-metrics}, we can observe the RMSE, MCE, and failure frequency are much higher in the \textit{Comma.ai} dataset compared to the \textit{Udacity} simulation. For instance, using ATN with ResNet, RMSE, MCE and Fails are 1.09, 0.27 and 2.3 in simulation compared to 3.12, 0.77 and 3.4 in the \textit{Comma.ai} dataset. The error nearly doubles when transferring from simulation to the real world. This implies the much higher complexity and diversity in real-world environments, which provides greater challenges. This suggests the need for future work in training and testing on a larger real-world dataset and more realistic simulation environment.

\section{Conclusion}
In this paper, we proposed a network architecture that improves the baseline vision-based End-to-End learning of steering angle with auxiliary tasks. The proposed network architecture considers five distinct auxiliary tasks: using auxiliary image segmentation, transferring from existing tasks, adding a recurrence module, introducing optical flow and incorporating vehicle kinematics. We find that the features learned from the Imagenet recognition task can be transferred and beneficial to the task of steering for on-road driving. The pre-trained segmentation mask to categorize the image at the pixel level can empower the network with more information and thus result in better prediction accuracy. The incorporation of temporal information with a recurrence module and optical flow helps to improve the current action choices. Finally, the proper addition of some vehicle kinematics makes the state representation more concrete and helps to boost performance.

\section*{Acknowledgment}
This work was funded by General Motors Global R\&D. The authors would like to thank Yihao Qian and Abhinav Gupta for fruitful discussions about the project.

{\small
\bibliographystyle{ieee}
\bibliography{egbib}
}

\end{document}